\documentclass{article}
\usepackage[nonatbib, preprint]{neurips_2022}

\usepackage[numbers]{natbib}

\usepackage{todonotes}
\usepackage[utf8]{inputenc} 
\usepackage[T1]{fontenc}    
\usepackage{hyperref}       
\usepackage{url}            
\usepackage{booktabs}       
\usepackage{amsfonts}       
\usepackage{nicefrac}       
\usepackage{microtype}      
\usepackage{graphicx}
\usepackage{multicol}
\usepackage{multirow}
\usepackage{longtable}
\usepackage{caption}
\usepackage{subcaption}
\usepackage{wrapfig}
\usepackage[titletoc]{appendix}

\usepackage[normalem]{ulem}
\newif{\ifhidecomments}

\makeatletter
\newcommand\footnoteref[1]{\protected@xdef\@thefnmark{\ref{#1}}\@footnotemark}
\makeatother
\title{[Re] Badder Seeds: Reproducing the Evaluation of Lexical Methods for Bias Measurement}
\newcommand*\samethanks[1][\value{footnote}]{\footnotemark[#1]} 
\author{%
  Jille van der Togt\thanks{Equal Contribution} \\
  University of Amsterdam\\
  Amsterdam, the Netherlands\\
  \texttt{jille.vandertogt@student.uva.nl} \\
   \And
  Lea Tiyavorabun\samethanks \\
  University of Amsterdam\\
  Amsterdam, the Netherlands\\
  \texttt{lea.tiyavorabun@student.uva.nl} \\
  \And
  Matteo Rosati\samethanks \\
  University of Amsterdam\\
  Amsterdam, the Netherlands\\
  \texttt{matteo.rosati@student.uva.nl} \\
  \And
  Giulio Starace\samethanks~~\thanks{Corresponding Author} \\
  University of Amsterdam\\
  Amsterdam, the Netherlands\\
  \texttt{giulio.starace@gmail.com}
}
\begin{document}
\raggedbottom
\maketitle
\section{Reproducibility Summary}

\subsection*{Scope of Reproducibility}
Combating bias in NLP requires bias measurement. Bias measurement is almost always achieved by using \emph{lexicons of seed terms}, i.e. sets of words specifying stereotypes or dimensions of interest. This reproducibility study focuses on \citet{antoniak-mimno-2021-bad}'s main claim that the rationale for the construction of these lexicons needs thorough checking before usage, as the seeds used for bias measurement can themselves exhibit biases. The study aims to evaluate the reproducibility of the quantitative and qualitative results presented in the paper and the conclusions drawn thereof.

\subsection*{Methodology}
We re-implement the entirety of the approaches outlined in the original paper. We train a skip-gram word2vec model with negative sampling to obtain embeddings for four corpora. This does not require particular computing requirements beyond standard consumer personal computers. Additional code details can be found in our linked repository.

\subsection*{Results}
We reproduce most of the results supporting the original authors' general claim: seed sets often suffer from biases that affect their performance as a baseline for bias metrics. Generally, our results mirror the original paper's. They are slightly different on select occasions, but not in ways that undermine the paper's general intent to show the fragility of seed sets.

\subsection*{What was difficult}
The significant difficulties encountered were due to a lack of publicly available code and documentation to clarify missing information in the paper. For this reason, many algorithms that ultimately turned out to be quite simple required lengthy clarifications with authors or trial and error. Lastly, the research was quite data-intensive, which caused some implementations to be non-trivial to account for memory management.

\subsection*{What was easy}
Once understood, the methods proposed by the authors were relatively easy to implement. The mathematics involved is quite straightforward. Communication was also reasonably accessible. The authors' emails were readily available, and the responses came quickly and were always helpful.

\subsection*{Communication with original authors}
We maintained a lengthy email correspondence throughout the replication of the paper with one author, Maria Antoniak. We contacted her to clarify extensive aspects of the paper's methodology. Specifically, this concerned summarizing the data processing approach,  explaining missing hyperparameters, and outlining the aggregation of metrics across different bootstrapped models. None of the original code was disclosed.

\clearpage

\section{Introduction}
The emergence of bias quantification in Natural Language Processing (NLP) methods has given rise to two use cases, referred to as \emph{downstream} and \emph{upstream}. In the former, bias measurements are used to debias or correct biases in word representations to avoid encoded biases trickling down when applying these NLP models \citep{bolukbasi_man_2016, caliskan_semantics_2017}. In the latter, bias measurements are used on models trained on small corpora to quantify the bias present and compare them.
This use case has endowed social scientists with the quantitative foundation to answer political and
social questions about bias across corpora in an empirical manner
\citep{rudinger_social_2017,joseph_girls_2017}.
Crucially, most bias quantification methods depend on lexicons of seed terms that specify the bias dimensions of interest. The selection of seed terms varies considerably across the literature, and seed sets themselves may exhibit social and cognitive biases \citep{antoniak-mimno-2021-bad}. It is not clear whether it is possible to re-use seed set across corpora (thereby interfering with \emph{upstream} use cases), and elements such as seed term frequency have been shown to affect bias measurements, and thus \emph{downstream} uses \citep{ethayarajh2019understanding}.

We seek to replicate the \citet{antoniak-mimno-2021-bad} paper, hereafter referred to as ''the original paper/work''. In it, the authors seek to 1) qualitatively explore seed selection and their sources, 2) demonstrate that features of seed sets such as pairing order, set similarity, and frequency can cause instability in bias measurements, and 3) make recommendations for the testing and justifying of seed sets in future work. We have replicated the experiments showing the fragility of seed sets, thus verifying the claims of a need for better justification and analysis of them in future literature. We have also built a public toolkit to reproduce these measures on arbitrary seed sets and trained embeddings.

\section{Scope of reproducibility} \label{sec:claims}
This reproducibility study focuses on the authors' main claim that seed lexicons need thorough checking before usage to measure bias, as seeds themselves can be biased and induce instabilities in measurement. The authors conducted a literature review on prior works to gather many seed sets. They subsequently evaluated the gathered seed sets with a series of bias measurement metrics proposed by \citet{bolukbasi_man_2016, caliskan_semantics_2017}, and themselves.

Our work consists of two interconnected efforts: code replication, given the absence of pre-existing code for the original paper, and reproducing the main results. The latter goal is the main focus of our work and entails reproducing the outcomes that support the paper's central claims, which can be summarized as follows:

\begin{enumerate}
	\item Bias subspaces generated from common bias subspace metrics (e.g., WEAT, PCA) can help capture the difference represented by the seed set pairs.\label{claim:subspace}
	\item Bias subspaces suffer from instability due to the following factors:
	      \begin{enumerate}\label{claim:instability}
		      \item The ordering and pairing of the seed sets.\label{claim:instability:ordering}
		      \item The selection of seeds that are members of the seed sets.\label{claim:instability:choice}
		      \item The degree of semantic similarity between seeds. \label{claim:instability:semantic}
	      \end{enumerate}
	\item Methods of sourcing seed sets are inconsistent, with disparate strategies being used across NLP literature. \label{claim:sources}
\end{enumerate}

\section{Methodology}

The code from the original paper was not made publicly available. We, therefore, re-implemented the
entire approach from the description in the original paper. The following section will summarize the
resources and methodology used to reproduce the original paper accurately.

\subsection{Code}

As mentioned above, the code from the original paper is not publicly available. We fully
re-implement all the code, which can be found on
GitHub\footnote{\href{https://github.com/thesofakillers/badder-seeds}{https://github.com/thesofakillers/badder-seeds}}.
We closely follow the original paper's methodology to achieve accurate reproduction. The
reproduction is performed step by step, from downloading and preprocessing the data to training the
models and visualizing the results.

\subsection{Documentation}

Unfortunately, there was little to no documentation in the original work besides the content of the
original paper. This occasionally lacked crucial information to reproduce the results or was vague
on implementation details. In addition to the original paper, \citet{antoniak-mimno-2021-bad}
published a Github repository that contained a JSON with the metadata on seed sets gathered from
prior
works\footnote{\label{footnote:badseedsrepo}\href{https://github.com/maria-antoniak/bad-seeds}{https://github.com/maria-antoniak/bad-seeds}}.

\subsection{Model descriptions}

We train several bootstrapped skip-gram word2vec models with negative sampling on unigrams on each
dataset. This model attempts to predict whether a particular word is a valid context (where the
context window size is a hyperparameter) for a given other word using a single fully connected
hidden layer. The first step in training this model is creating a vocabulary of the entire training
dataset. With this vocabulary, each word can be represented as a one-hot vector. The network output
is then a measure of the probability that the word is a valid context. The trained weights from this
hidden layer are then used to obtain word embedding vectors for each term in the training set
vocabulary.

\subsection{Datasets}

The original paper used four datasets and one pretrained model: New York Times articles from April
15th-June 30th,
2016\footnote{\href{https://www.kaggle.com/nzalake52/new-york-times-articles}{https://www.kaggle.com/nzalake52/new-york-times-articles}};
high-quality WikiText articles, using the complete WikiText-103 training set
\citep{DBLP:journals/corr/MerityXBS16}; Goodreads book reviews for the romance and history and
biography genres sampled from the UCSD book Graph \citep{10.1145/3240323.3240369,
	wan-etal-2019-fine}; and the pretrained word2vec GoogleNews
model\footnote{\label{footnote:googlenews}\href{https://github.com/mmihaltz/word2vec-GoogleNews-vectors}{https://github.com/mmihaltz/word2vec-GoogleNews-vectors}}.
We use these same corpora for our research, preprocessing them as closely as possible to the
original paper. This consists of grouping the text into documents, filtering relevant documents,
lowercasing and removing special characters. We then use spaCy
\citep{honnibalSpaCyIndustrialstrengthNatural2020} for tokenization and POS-tagging. Because the
work is not concerned with model performance, this study makes no use of train/dev/test splits. The
WikiText-103 dataset, however, is pre-split, so like in the original work, we work with the training
split. Links to all these datasets can be found in our Github repository.

Preprocessing statistics of our work and the original paper can be found in Table
\ref{tab:tab_2_repl}. We find general agreement in our numbers regarding the total number of
documents per dataset. There are minor discrepancies in the Goodreads datasets, most likely due to
implementation differences. We also count slightly fewer total words than the original paper in all
cases, but the orders of magnitude generally match. We are, however, unable to reproduce vocabulary
size accurately. We tried many strategies in the replication process to obtain these numbers, but
none were successful. Furthermore, looking at the official dataset statistics, for example for
WikiText \citep{DBLP:journals/corr/MerityXBS16}, it is clear that our reproduced vocabulary size is
a lot closer to the ground truth than the one by \citet{antoniak-mimno-2021-bad}. Lastly, mean
document length values of each dataset are accurately reproduced, with the WikiText values suffering
the most. The subsections below will discuss each dataset in more detail.

\paragraph{New York Times} This dataset contains 165,900 paragraphs from 8,888 articles from the New
York Times published between April 15th and June 30th 2016. The articles cover a broad range of
sections, including but not limited to movies, sports, technology, business, books, science, and
fashion.

\paragraph{WikiText-103} This dataset contains 28,472 manually verified articles from Wikipedia.org.
The entire training dataset is used, in which lists, HTML errors, math, and code have already been
removed. Furthermore, we removed all formulas still present in the text.

\paragraph{Goodreads} The entire Goodreads dataset contains millions of reviews. This study uses
just the Romance and the History/Biography genres. Five hundred book reviews per book are sampled
for each genre while filtering out all books with fewer than 500 reviews and all reviews containing
fewer than 20 characters.

\paragraph{GoogleNews} Google's pretrained word2vec model is trained on ca.100 billion words from
the GoogleNews dataset (\ref{footnote:googlenews}). Our use of this model was limited to replicating
the results outlined below for additional robustness.

\paragraph{Seed Set Dataset} Part of the contributions of the original work was creating a catalogue
of 178 seed sets gathered from eighteen highly-cited prior works on bias measurements. We refer to
this catalogue as the \emph{gathered seeds}. Each element of the catalogue comprises a seed set, the
category it represents, a justification, the source categorization, a link, and a unique ID. It is
readily available on the original author's GitHub\footnoteref{footnote:badseedsrepo}. A brief
statistical overview can be found in Fig. \ref{fig:repro_fig1}. We process the catalogue by
lower-casing the seeds and removing bigrams to use them with our models. We also filter seed sets
containing less than two seeds as we argue that a single seed would not be sufficient to form a set.

\subsection{\label{sec:calc_agg}Experimental setup and code} An environment containing all necessary
packages is included in the publicly available repository and can be quickly set up. To mirror the
original paper's setup, we used the \emph{gensim} \citep{rehurek2010software} implementation of
skip-gram with negative sampling \citep{mikolov2013distributed} to train the vector embeddings for
all datasets. We used this library to train our models as that is the framework used by the original
paper and to avoid noise due to different implementations (the investigation of which would be
outside the scope of this paper). Several PyTorch \citep{pytorch} implementations are also available
on GitHub if that is
preferred\footnote{\href{https://github.com/theeluwin/pytorch-sgns}{https://github.com/theeluwin/pytorch-sgns}}\textsuperscript{,}\footnote{\href{https://github.com/ddehueck/skip-gram-negative-sampling}{https://github.com/ddehueck/skip-gram-negative-sampling}}.

We reproduce the original paper's results by focusing on two popular seed-based bias metrics to
measure bias in corpus-derived embeddings: WEAT and PCA. These metrics are used to produce
a \emph{bias subspace} vector given a pair of seed sets that specifies a bias dimension of interest.
The WEAT method, introduced in \citet{caliskan_semantics_2017}, produces a vector based on the
difference between the mean vectors of the two target sets. The PCA method, described in
\citet{bolukbasi_man_2016}, instead requires that each seed term in one of the seed sets be paired
with one seed term from the other seed set. The subspace vector is then the first principal
component resulting from the PCA of a matrix constructed by, for each pair of seeds, taking the two
half vectors from the pair's mean to the two pair members and using them as two columns of the
matrix.

We also reproduce the original paper's coherence metric, which aims to quantify the robustness of
the bias subspace. This metric is calculated as the absolute value of the difference in mean ranks
of the terms in two seed sets when all the model's vocabulary is ranked by cosine similarity to the
bias subspace. Another metric used is set similarity, the cosine similarity between the average
vectors of two seed sets.

Finally, when aggregating embeddings of a specific word across bootstrapped models, we take the
average of the embedding vectors in each model that includes the word. Given a particular pair of
seed sets for coherence aggregation, we only average coherence scores for models containing every
seed term in the two sets to avoid aggregating coherence based on different seed sets.

\subsection{Hyperparameters} 100-dimensional embeddings were trained for five epochs on all four
datasets, with a five-word negative context sampling rate and a window size of five. We trained
embeddings with a minimum word count of 0, 10, and 100 due to variation in the original paper. This
process was repeated for 20 bootstrapped samples of each dataset (with the sample size equal to the
number of documents in the dataset), resulting in 20 separate models. The bootstrapping provided the
stochasticity required for robustness. To ensure this reproducibility, we use a random seed of 42
throughout.

\subsection{Computational requirements} The execution of the reproduced code does not take excessive
computing power. This study used no GPUs or computing clusters. We ran the experiments on an Intel
I9 9900k and 32GB of 3200MHz RAM running Ubuntu 20.04.3 LTS. Table \ref{tab:compute} shows peak RAM
usage and time in seconds to completion for every subprocess of the replication.

\section{Results} \label{sec:results}

\subsection{Quantitative Results} 
\begin{figure}[ht]
	\centering
	\begin{subfigure}[b]{0.6\textwidth}
		\centering
		\includegraphics[scale=0.55]{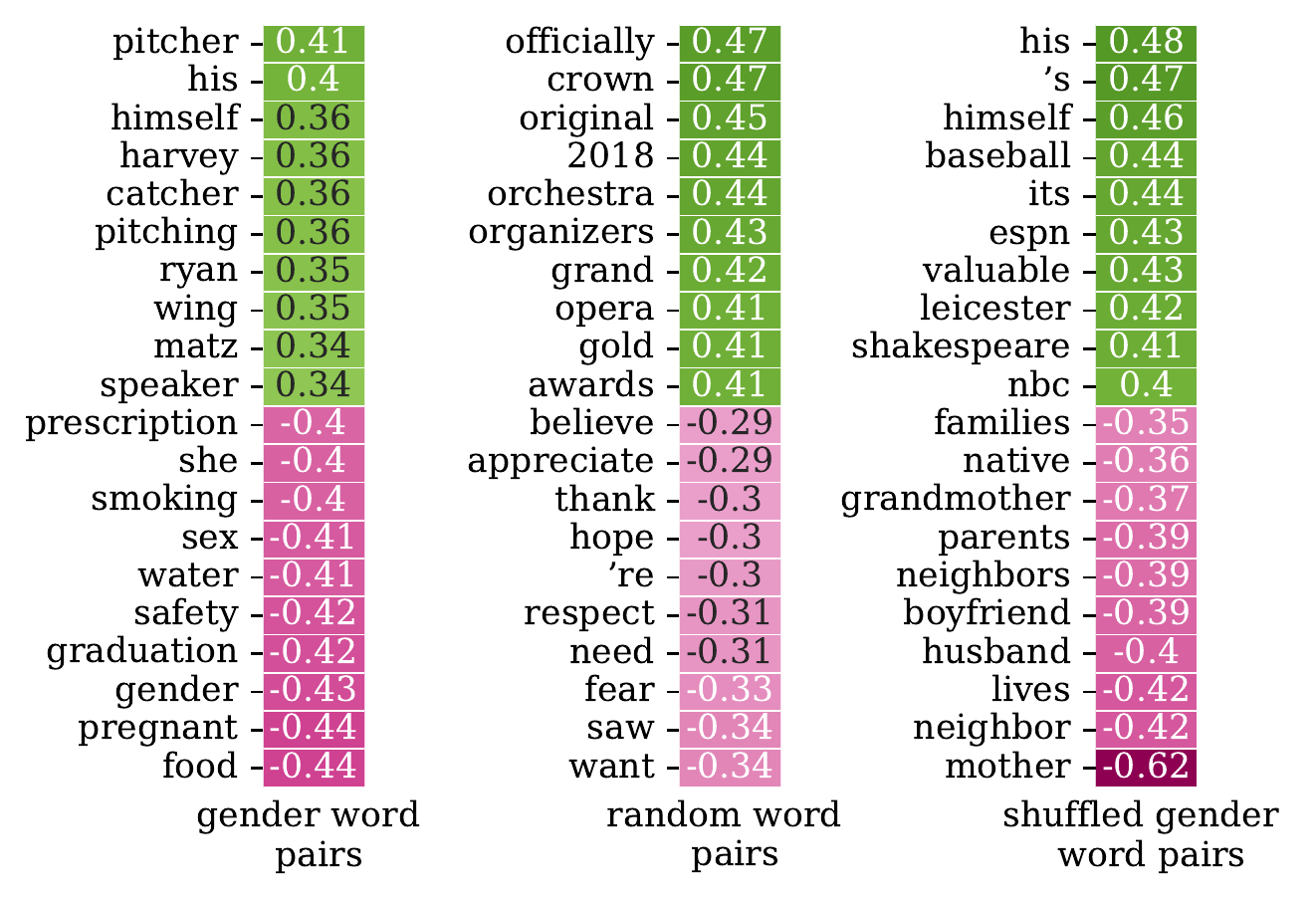}
		\caption{Compares the top and bottom ten words of each bias subspace ranked by cosine similarity out of all words in the corpus.}
		\label{fig:repro_4_1}
	\end{subfigure}
	\hfill
	\begin{subfigure}[b]{0.33\textwidth}
		\centering
		\includegraphics[scale=0.46]{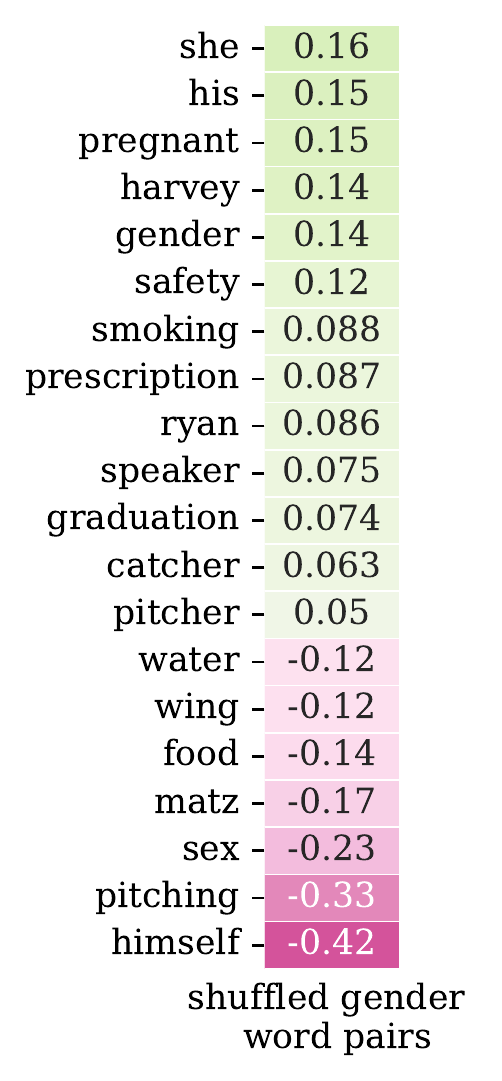}
		\caption{Top \& bottom ten words of ordered subspace ranked for the shuffled subspace.}
		\label{fig:repro_4_2}
	\end{subfigure}
	\caption{Replication of Fig. 4 of the original paper. Ranks words from corpus by cosine similarity against different bias subspaces (first principal component), with NYT frequency threshold 100.}
	\label{fig:repro_4}
\end{figure}

We started by confirming that the bias subspace does capture the
difference or bias that the seed pairs are intended to represent. For this, we reproduced an
experiment by \citet{antoniak-mimno-2021-bad} ranking the cosine similarity between the first
Principal Component (PC) of the bias subspace and all words in the corpus. The top and bottom ten
words for each bias subspace are shown in Fig. \ref{fig:repro_4_1}. In the shown words of the
\textit{gender pair subspace} and the \textit{shuffled gender pair subspace} gender-related words
are found, whereas none are present in the \textit{random pair subspace}. However, only the
\textit{gender pair subspace} divides nicely between \textit{male} and \textit{female} terms. We
extended this by calculating the cosine similarity of the top and bottom ten words from the
\textit{ordered bias subspace} for the \textit{shuffled bias subspace}. The results in Fig.
\ref{fig:repro_4_3} show \textit{she} and \textit{his} as the two highest-ranked words, which are
not split along the intended bias subspace.

\begin{figure}[ht]
	\centering
	\includegraphics[scale=0.5]{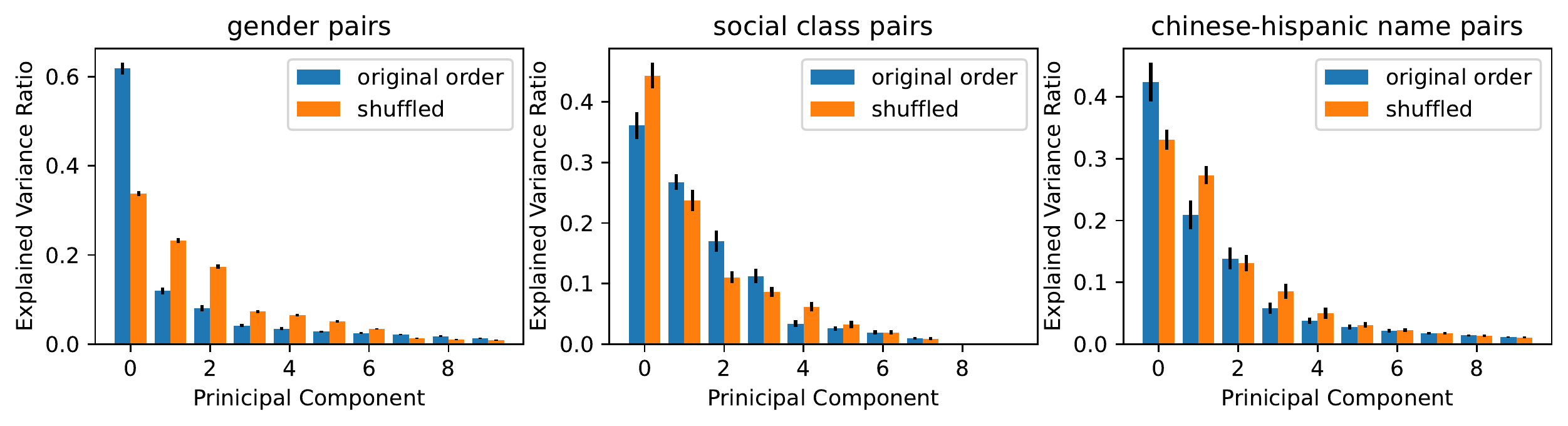}
	\caption{Replication of Fig. 3. The first ten principal components of the bias subspace for different seed pairs on the NYT corpus with a minimum frequency of 0.}
	\label{fig:repro_3}
\end{figure}

Fig. \ref{fig:repro_3} shows that the first PC has almost always a very high explained variance
ratio for the bias subspace of ordered pairs, which drops off quickly for the subsequent PCs.
Instead, the explained variance ratio per PC drops more smoothly for the shuffled pairs. Fig.
\ref{fig:repro_4_3} shows this behavior by computing the top and bottom ten words by cosine
similarity against the second PC of the gender subspace. We can observe that the bias subspace of
the ordered pairs does not contain gender words anymore. In contrast, the shuffled subspace does
have gender words such as \textit{her}, thereby replicating the trend observed in Fig.
\ref{fig:repro_3}. It is also important to note that in Fig. \ref{fig:repro_3} there are exceptional
cases where shuffled seed sets produce the first PC with a higher explained variance than the
ordered seed sets. In general, these results replicate the trends of the original experiments.

\begin{wrapfigure}{R}[5pt]{0.7\textwidth}
	\centering
	\includegraphics[width = 0.65\textwidth]{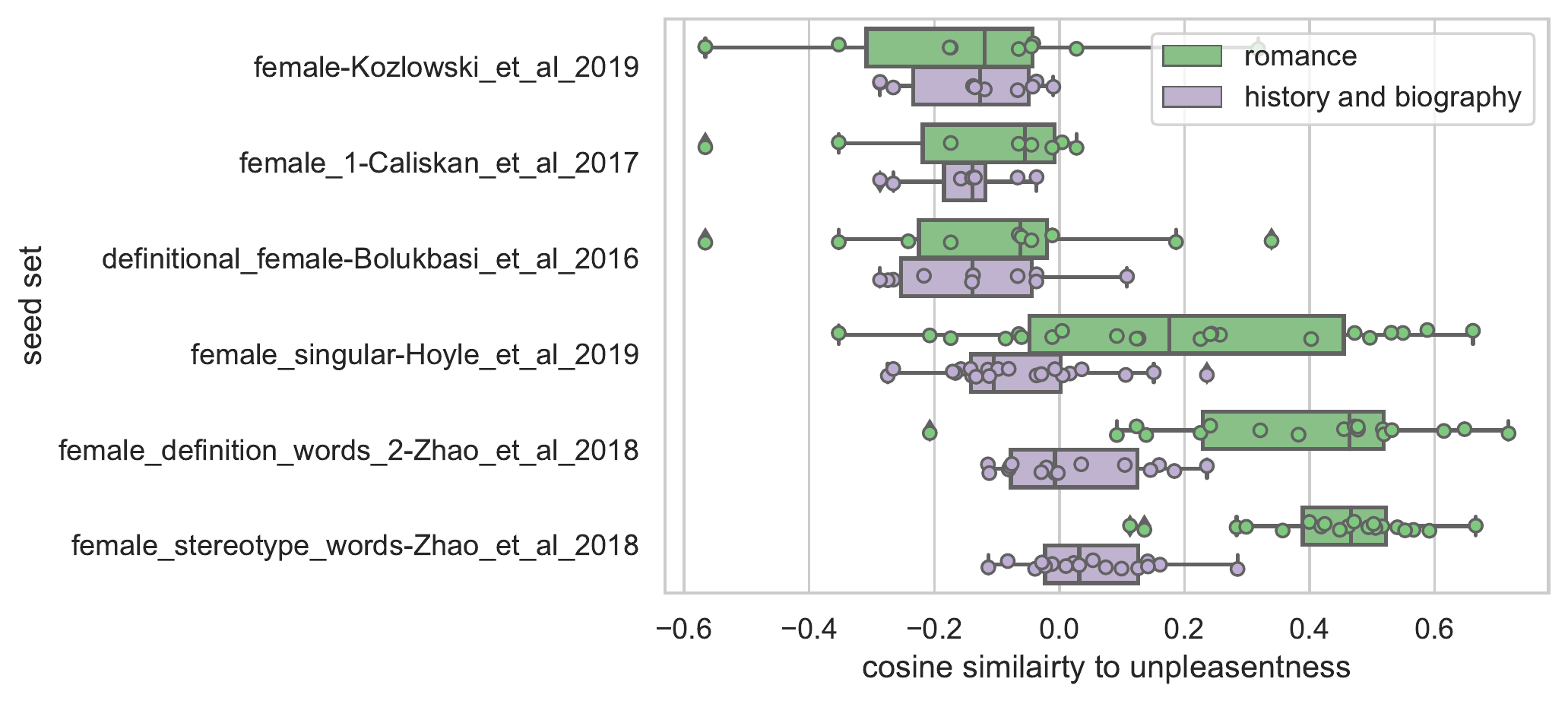}
	\caption{\label{fig:repro_2}Reproduction of Fig. 2. Displaying the cosine similarity between the averaged vector of \textit{unpleasantness} across all 20 bootstrapped models and different seeds sets of the category \textit{female}.}
\end{wrapfigure}

Fig \ref{fig:repro_2} shows that bias measurement is highly inconsistent across seed sets with the
same seed category sourced from different papers. We used the cosine similarity between
\textit{female} seed sets and the word \textit{unpleasantness} as a bias measurement. The cosine
similarity varies greatly between seed sets, replicating the same trends as the original paper.

\begin{figure}[ht]
	\centering
	\includegraphics[scale=0.39]{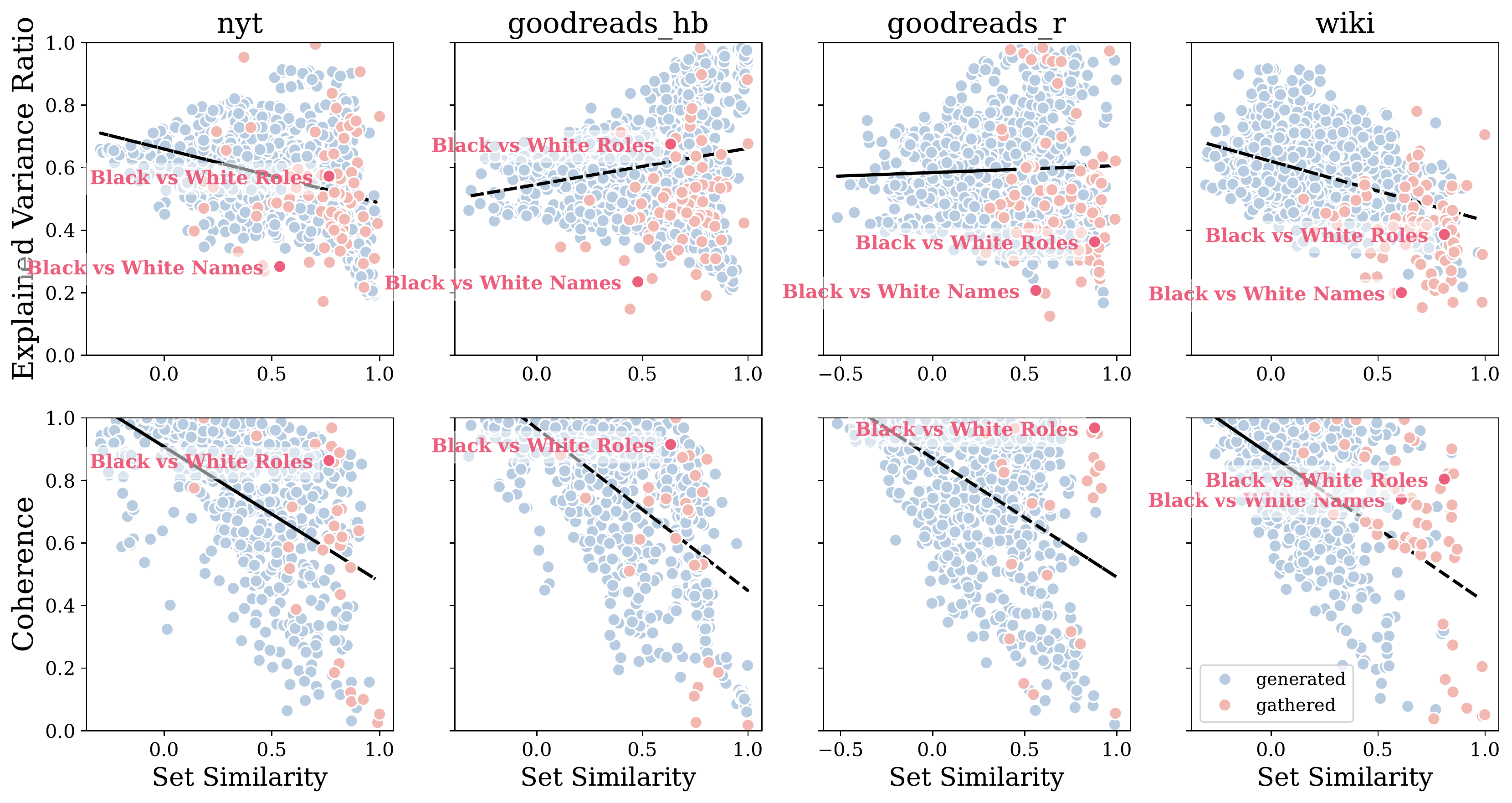}
	\caption{\label{fig:repro_5}Replication of Fig. 5 from the original paper, displaying Explained Variance Ratio (top) and Coherence (bottom) vs Set Similarity across the four datasets. We highlight two pairs of gathered seed sets, Black vs White roles and names. For some corpora, seed terms were not found in the embeddings, causing the highlighted pair to be missing.}
\end{figure}

Fig. \ref{fig:repro_5} explores the relationship between set similarity and the robustness of the
bias subspace. The relationship between set similarity and the explained variance of the PCA-derived
bias subspace vector is plotted for each dataset and frequency thresholds. The original paper shows
this relationship only for the WikiText dataset, and we find a similar negative correlation between
set similarity and explained variance for that dataset.

Table \ref{tab:coherence} qualitatively explores this relationship, ranking both gathered and
generated sets by coherence. More semantically dissimilar seed sets score higher in coherence than
more similar sets. In the gathered sets, seed sets related to names have extremely low coherence due
to their semantics being very similar and the set pairs containing duplicate terms (see "names
black" and "names white"). In the generated sets, we see that very different terms (such as those
relating to careers and those related to lower body clothing/parts) have high coherence. In
contrast, sets such as food terms score much lower. We observe a similar pattern when using the PCA
algorithm as a basis for coherence. These results show the replicability of the original paper, as
they are almost identical.

\subsection{Qualitative Results} The original paper gathered 178 seed sets of eighteen highly-cited
prior work on bias measurement. These seeds are both embedding-based and non-embedding-based bias
detection methods, often overlapping. The seeds are chosen in a multitude of ways. Only unigram
seeds are selected, and words that do not appear in the training corpus are omitted. We have
validated the accuracy of Table 3 in the original paper by reviewing each of the eighteen papers and
determining which methods the authors used. We briefly summarize them below:

\paragraph{Borrowed from social sciences} Select seed sets are borrowed from prior psychology and
other social sciences work.
\paragraph{Crowd-Sourced} Crowd-based annotation can create custom seed sets. This method can aid in
gathering contemporary associations and stereotypes. However, controlling crowd demographics often
poses a problem. This can lead to stereotypes being hard-coded into the seeds.
\paragraph{Population-Derived} Seeds can also be derived from government-collected population
datasets. These datasets are usually names and occupations common to specific demographic groups.
A significant problem with this method is that the data tends to be often US-centric and thus gives
a distorted view of the rest of the world.
\paragraph{Adapted from Lexical Resources} Researchers can also draw seeds from existing
dictionaries, lexicons and other public resources. The advantage is that these seeds have already
undergone a round of validation.
\paragraph{Corpus-Derived} This quantitative method is used to extract seeds terms from a corpus. It
has the advantage of ensuring high-frequency words are selected but suffers from similar risks as
crowd-sourced seeds.
\paragraph{Curated} Seed hand-selection by authors often yields high precision seeds but is slow and
relies on unbiased authors.
\paragraph{Re-used} The last method relies on prior bias measurement research for seed terms. The
advantage is that the seeds have already been used, but researchers should not use them without
validation.

\subsection{Results beyond original paper}

\paragraph{Set Similarity and Bias Subspace in Additional Datasets} We extended the original paper's
set similarity versus bias subspace explained variance analysis to cover all datasets (beyond
WikiText) in Fig. \ref{fig:repro_5}. The negative trend is still present with the NYT corpus, but
not in the Goodreads corpora, where the trend is almost absent or slightly positive. In addition,
the positions of the highlighted seed set pairs are variable across corpora. We also extended this
work to examine the relationship between seed pair coherence versus set similarity, where the
inverse relationship is present in all datasets. Notice that the requirement that coherence is
calculated only for models that contain all seed terms (as described in Section \ref{sec:calc_agg})
makes specific pairs of seed sets be ignored, as seen from the lack of the two highlighted set pairs
for select datasets.

\paragraph{Testing Minimum Frequency Filter} Due to inconsistencies both in the paper and in
communication with the author in the reported minimum frequency filter for the skip-gram models, we
experimented with minimum frequencies $\mu \in \{ 0, 10, 100\}$. These enabled us to see results
across the whole vocabulary in the case of $\mu = 0$ and reduce noise from rare words in the case of
$\mu = 10$. We also used $\mu = 100$ to generate Fig. \ref{fig:repro_4} as the original paper.

\paragraph{Seed Toolkit and Pairing Seed Set Data.} Other than extending the experiments of the
original paper, we have two additional contributions. For the sake of reproducibility, we make our
code publicly available and design our repository as an open Python package that can be used to
obtain bias subspace vectors and assess seed set robustness. This toolkit can help future
researchers who aim to evaluate their seeds carefully. Our second contribution is an augmentation of
the seed dataset provided by \citet{antoniak-mimno-2021-bad}. We provided additional annotations
regarding pairing, i.e. we identify which seeds to pair together along standard bias dimensions in
a queriable \texttt{.csv} format.

\section{Discussion}

Overall, our results replicate the data reported in the original paper. This replication lends
strong support to the general claim of the original paper that seed sets incorporate strong
inductive biases that affect their performance as grounding for bias metrics and that researchers
should be more cognizant of these limitations.

Instability in bias subspaces can be introduced by selecting seeds in seed sets, as stated in claim
\ref{claim:instability:choice}. Our results in Fig. \ref{fig:repro_2} support this as they reproduce
the original work. The same bias measurement varies across seed sets selected by different authors
who assigned it to the same category. In addition, the dependence of the bias subspace on seed set
selection is further supported by Fig. \ref{fig:repro_5}. The two highlighted seed sets (black vs
white roles/names) are generally distinct in position for each corpus, despite theoretically
attempting to define similar bias dimensions.

Another source of instability claimed by \citet{antoniak-mimno-2021-bad} is the ordering and pairing
of seed sets. In Fig. \ref{fig:repro_3} we show that the explained variance ratio for the
\textit{ordered bias subspaces} can behave very differently from the \textit{shuffled bias
	subspaces}, supporting claim \ref{claim:instability:ordering}. Our work in Fig. \ref{fig:repro_4_1}
also supports this claim. While the ordered subspace successfully splits the top words along the
intended subspace of \textit{male} and \textit{female}, the first PC of the shuffled bias subspace
has words such as \textit{mother} and \textit{boyfriend} both ranked on the same end. This shows
that while the subspace still picks up on gender words, it does not represent the intended subspace.
Supporting claim \ref{claim:instability:ordering} that bias subspaces can become less meaningful
with a shuffled seed pairing. We could further confirm this behavior by calculating the cosine
similarity of the top words of the \textit{ordered subspace} for the \textit{shuffled subspace} in
Fig. \ref{fig:repro_4_2}. These results show that \textit{she} and \textit{his} are ranked next to
each other at the top and not split along the intended bias subspace. These experiments lend strong
support to claim \ref{claim:instability:ordering} that the order of seed pairs can substantially
influence the meaningfulness of the bias subspace and, consequently, the bias metrics.

Finally, bias subspaces suffer instability due to semantically overlapping seeds being less
distinguishable in the bias subspace, as stated in claim \ref{claim:instability:semantic}. Our
results in Table \ref{tab:coherence} and Fig. \ref{fig:repro_5} demonstrate that bias subspace
vectors are less robust when the seed sets are semantically similar or overlapping. This
relationship lends strong credence to claim \ref{claim:instability:semantic}. However, our results
did show that this inverse relationship is not conserved across a minority of corpora (e.g., the
Goodreads datasets) for the explained variance metric. More broadly, however, this still shows that
the reliability of seed selection is quite variable. While similar seed sets may generate robust
bias subspaces for more semantically equivalent seed pairs for some corpora, that is not guaranteed.
Therefore, while this inverse relationship may be minimized for specific corpora, extensive
corpus-specific seed set investigations are still required.

\paragraph{What was easy.} The original paper clearly described the algorithms used to obtain bias
metrics. Additionally, it carefully cited the papers that first proposed them, which specified
further details. This aided our understanding of the underlying concepts and accelerated the
implementation of the frameworks. Model training and embedding generation was also facilitated by
the pre-existing \emph{gensim} framework. This permitted greater focus on reproducing the details of
the experiments than choosing between alternative implementations of skip-gram word2vec. In
addition, responsive authors permitted quick clarifications through email communication when
important details were not clear.

\paragraph{What was difficult.} The original paper did not make code publicly available and largely
lacked documentation. Only the gathered seeds were provided via GitHub
(\ref{footnote:badseedsrepo}). This made it necessary to reproduce all the code from scratch.

In select instances, the paper crucially omitted important information, making us reliant on
communication with the authors. This was most pronounced when aggregating embeddings or other
metrics across the bootstrapped model sampling, where vocabulary sizes were different. This meant
that not all models had good embeddings for all seed terms. We had to consider several different
approaches before settling on the averaging criteria described in Section \ref{sec:calc_agg}.

Finally, preprocessing the data was more difficult than initially imagined. The tokenization
pipeline in the original paper was vaguely specified, and differences in our implementation caused
the slight discrepancies in Table \ref{tab:compute}. The POS tagging with spaCy was imperfect,
resulting in the incorrect tagging of several proper nouns as common nouns, making it hard to
control for POS in random seed generation.

\paragraph{Communication with original authors.} While the authors did not disclose any code, we
maintained a lengthy email correspondence with them. One author, Maria Antoniak, was contacted to
clarify hyperparameters of the word2vec model, the methodology for generating random seeds across
bootstrapped models, and which bias metrics (PCA or WEAT) were used for different results. She also
described her dataset processing pipeline, as there were many alternate ways to process the corpora
before training.

\section{Conclusion}

Overall, our results replicate the ones reported in the original paper. This lends strong support to
the general claim of the original paper that seed sets incorporate significant inductive biases that
affect their performance as grounding for bias metrics and that researchers should be more cognizant
of these limitations. Aside from confirming the danger of blindly using seed sets, we also provide
additional contributions. First of all, all code used to replicate the original paper is publicly
available. This code can obtain bias subspace vectors and assess seed set robustness. Secondly, we
extended the original paper's set similarity versus bias subspace explained variance analysis to
cover all datasets. Furthermore, we implement multiple numbers of minimum frequencies that further
enable results across the entire vocabulary. Lastly, we provide an additional annotation pairing of
the original seed dataset.

We have highlighted a need for carefully justifying the use of particular sets through empirical
means, but a theoretically sound and systematic method for doing so is still in its infancy. Further
work may explore what criteria seed sets should satisfy to demonstrate robustness. In addition,
future researchers may want to extend this work to bigram seed terms and embeddings to explore the
limitations of more expressive seeds and bias dimensions.
\newpage
\bibliography{references.bib}
\bibliographystyle{abbrvnat}
\newpage
\appendix
\section{Appendix}
\setcounter{figure}{0}
\renewcommand\thefigure{\thesection.\arabic{figure}}
\renewcommand\thetable{\thesection.\arabic{table}}

\begin{figure}[ht]
	\centering
	\includegraphics[scale = 0.5]{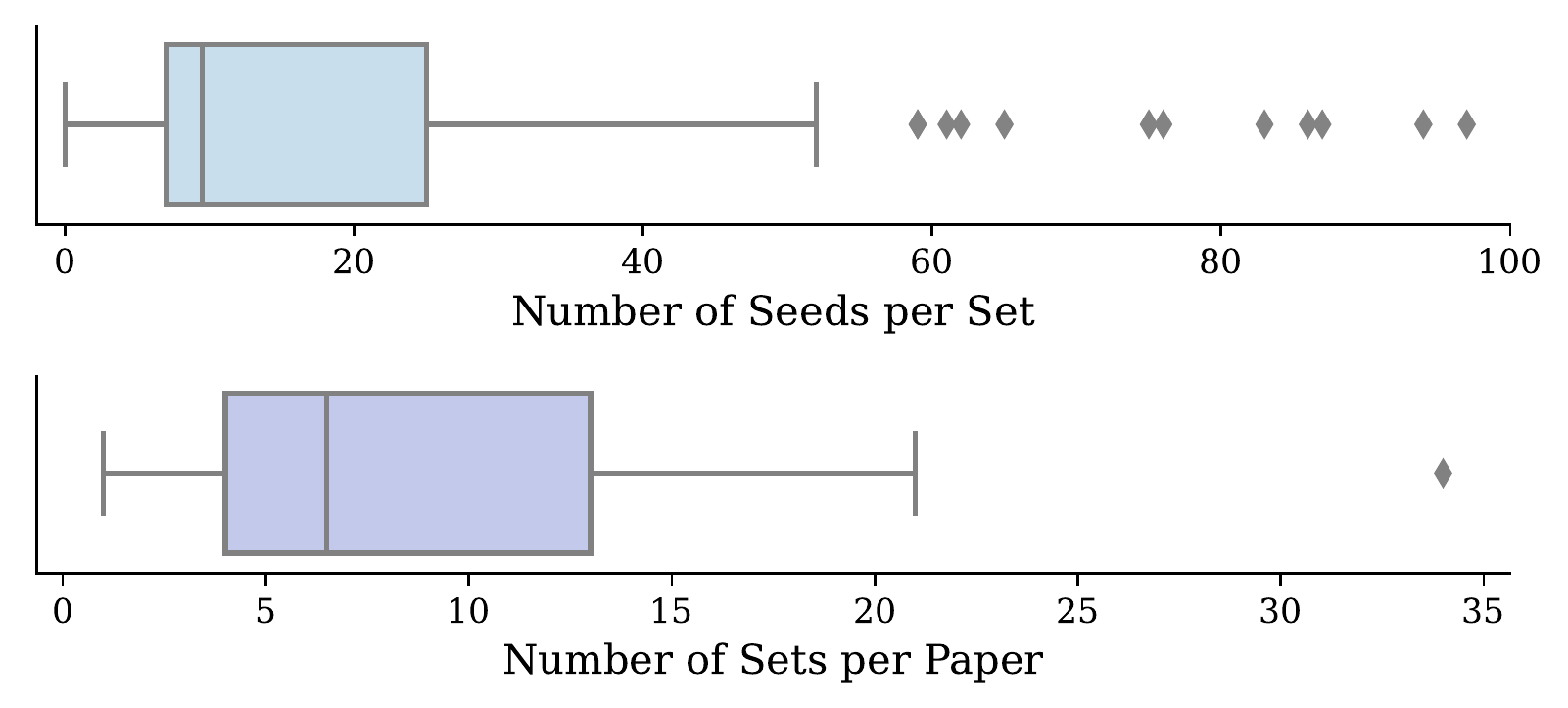}
	\caption{\label{fig:repro_fig1} Replication of Fig. 1 from the original paper, illustrating basic statistics of the gathered seeds.}
\end{figure}

\begin{table}[ht]
	\centering
	\caption{Comparing corpora summary statistics after preprocessing (original paper statistics obtained from Table 2). \label{tab:tab_2_repl}\newline}
	\resizebox{\columnwidth}{!}{%
		\begin{tabular}{@{}lllllllll@{}}
			\toprule
			\textbf{Dataset}                                                                       &
			\multicolumn{2}{l}{\textbf{\begin{tabular}[c]{@{}l@{}}Total\\ Documents\end{tabular}}} &
			\multicolumn{2}{l}{\textbf{Total Words}}                                               &
			\multicolumn{2}{l}{\textbf{Vocabulary Size}}                                           &
			\multicolumn{2}{l}{\textbf{\begin{tabular}[c]{@{}l@{}}Mean Document\\ Length\end{tabular}}}                                                                                                                                            \\ \midrule
			\textbf{}                                                                              & \textbf{original} & \textbf{ours} & \textbf{original} & \textbf{ours} & \textbf{original} & \textbf{ours} & \textbf{original} & \textbf{ours} \\
			\textbf{NYT}                                                                           & 8,888             & 8,888         & 7,244,457         & 7,217,851     & 162,998           & 109,713       & 815               & 812           \\
			\textbf{WikiText}                                                                      & 28,472            & 28,472        & 99,197,146        & 87,077,718    & 546,828           & 228,318       & 3,484             & 3,058         \\
			\textbf{Goodreads (Romance)}                                                           & 197,000           & 194,500       & 24,856,924        & 24,695,141    & 214,572           & 249,114       & 126               & 127           \\
			\textbf{Goodreads (History/Biog)}                                                      & 136,000           & 135,000       & 14,324,947        & 14,168,742    & 163,171           & 193,012       & 105               & 105           \\
			\bottomrule
		\end{tabular}%
	}
\end{table}

\begin{table}[ht]
	\caption{\label{tab:coherence} Replication of Table 4 from the original paper. Seeds that are more
		semantically similar have lower coherence scores. We use the WEAT metric (the difference between the
		mean vectors of the seed sets) to generate the subspace and the NYT dataset embeddings for this
		data. We average coherence scores across the $n$ models (out of $20$) that contain the paired seed
		sets and round to 3 decimal places. Unfortunately, while we tried to limit generated sets to only
		common nouns, proper nouns and, more rarely, verbs appeared in the sets due to issues with the spaCy
		POS tagger.\newline}
	\centering
	\resizebox{\columnwidth}{!}{%
		\begin{tabular}{lrl}
			\toprule
			\textbf{Coherence} & \textbf{Generated Set A}                              & \textbf{Generated Set B}                        \\
			\midrule
			1.000              & know, believe, think, guess, mean                     & governor, mayor, legislature, senator, democrat \\
			1.000              & foot-8, foot-7, foot-3, foot-5, to-4                  & rousteing, atkins, cornejo, ehrenreich, yorke   \\
			0.999              & associate, assistant, economist, engineer, accountant & heels, shoes, pants, legs, fingers              \\
			...                & ...                                                   & ...                                             \\
			0.062              & hertl, agnieszka, goran, brouwer, koivu               & bases, wings, outs, scoreless, rockies          \\
			0.059              & molina, glasser, pitney, darren, mackenzie            & carver, mina, boyce, curator, deputy            \\
			0.053              & lime, juice, lemon, potato, garlic                    & combo, bodysuit, raisin, koji, mango            \\
			\\
			\textbf{Coherence} & \textbf{Gathered Set A}                               & \textbf{Gathered Set B}                         \\
			\midrule
			0.999              & CAREER: executive, management, professio...           & FAMILY: home, parents, children, famil...       \\
			0.968              & MALE: brother, father, uncle, grandfat...             & FEMALE: sister, mother, aunt, grandmot...       \\
			0.942              & TERRORISM: terror, terrorism, violence,...            & OCCUPATIONS: banker, carpenter, doctor,...      \\
			...                & ...                                                   & ...                                             \\
			0.093              & MALE NAMES: john, paul, mike, kevin, ...              & FEMALE NAMES: amy, joan, lisa, sarah,...        \\
			0.053              & NAMES BLACK: harris, robinson, howard, ...            & NAMES WHITE: harris, nelson, robinson, ...      \\
			0.026              & NAMES ASIAN: cho, wong, tang, huang, ...              & NAMES CHINESE: chung, liu, wong, huang...       \\
		\end{tabular}%
	}
\end{table}

\begin{table}[htb]
	\centering
	\caption{Computing power needed for each action in the replication process.}
	\begin{tabular}{c|c|c}
		Action                     & Time (s) & RAM (MB) \\
		\hline
		Downloading the data       & 293      & 427      \\
		Preprocessing the data     & 3054     & 19018    \\
		Training all models        & 7806     & 21054    \\
		Table \ref{tab:tab_2_repl} & 4274     & 661      \\
		Fig. \ref{fig:repro_4}     & 22       & 4363     \\
		Fig. \ref{fig:repro_3}     & 19       & 4370     \\
		Fig. \ref{fig:repro_2}     & 4        & 1510     \\
		Fig. \ref{fig:repro_5}     & 500      & 1610     \\
	\end{tabular}
	\label{tab:compute}
\end{table}

\begin{figure}[htb]
	\centering
	\includegraphics[scale=0.8]{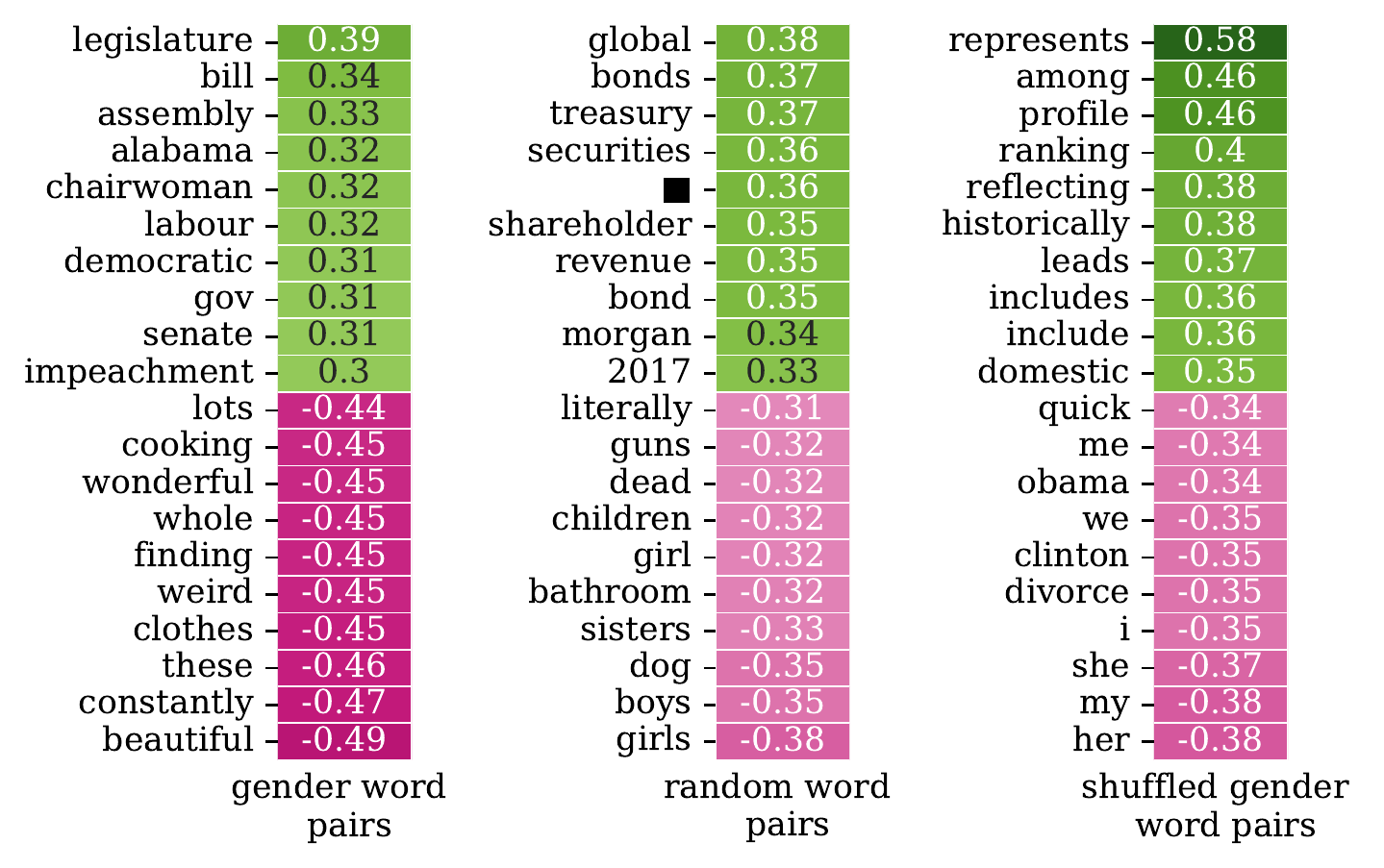}
	\caption{Extension on Fig. 4 from the original paper. Ranks words from the NYT corpus by cosine
		similarity against different bias subspaces (2nd principal component), with NYT frequency
		threshold 100. }
	\label{fig:repro_4_3}
\end{figure}

\begin{longtable}[htb]{p{0.1\linewidth}p{0.3\linewidth}p{0.6\linewidth}}
	\caption{Seeds used in various Figures\label{tab:seeds}}                                                                                                                                                                                                                                                           \\
	\toprule
	Figures                                               &
	Seed ID                                               &
	Seeds                                                                                                                                                                                                                                                                                                              \\ \midrule
	\multirow{6}{*}{Fig. \ref{fig:repro_2}}               &
	female-Kozlowski\_et\_al\_2019                        &
	{[}``woman", ``women", ``she", ``her", ``her", ``hers", ``girl", ``girls", ``female", ``feminine"{]}                                                                                                                                                                                                               \\
	                                                      &
	female\_1-Caliskan\_et\_al\_2017                      &
	{[}``sister", ``female", ``woman", ``girl", ``daughter", ``she", ``hers", ``her"{]}                                                                                                                                                                                                                                \\
	                                                      &
	definitional\_female-Bolukbasi\_et\_al\_2016          &
	{[}``woman", ``girl", ``she", ``mother", ``daughter", ``gal", ``female", ``her", ``herself", ``mary"{]}                                                                                                                                                                                                            \\
	                                                      &
	female\_singular-Hoyle\_et\_al\_2019                  &
	{[}``woman", ``girl", ``mother", ``daughter", ``sister", ``wife", ``aunt", ``niece", ``empress", ``queen", ``princess", ``duchess", ``lady", ``dame", ``waitress", ``actress", ``goddess", ``policewoman", ``postwoman", ``heroine", ``witch", ``stewardess", ``she"{]}                                            \\
	                                                      &
	female\_definition\_words\_2-Zhao\_et\_al\_2018       &
	{[}``lady", ``saleswoman", ``noblewoman", ``hostess", ``coquette", ``nun", ``heroine", ``actress", ``chairwoman", ``businesswoman", ``spokeswoman", ``waitress", ``councilwoman", ``stateswoman", ``policewoman", ``countrywomen", ``horsewoman", ``headmistress", ``governess", ``widow", ``witch", ``fiancee"{]} \\
	                                                      &
	female\_stereotype\_words-Zhao\_et\_al\_2018          &
	{[}``baker", ``counselor", ``nanny", ``librarians", ``socialite", ``assistant", ``tailor", ``dancer", ``hairdresser", ``cashier", ``secretary", ``clerk", ``stenographer", ``optometrist", ``housekeeper", ``bookkeeper", ``homemaker", ``nurse", ``stylist", ``receptionist"{]}                                   \\ \midrule
	\multirow{12}{*}{Fig. \ref{fig:repro_3}}              &
	definitional\_female-Bolukbasi\_et\_al\_2016          &
	{[}``woman", ``girl", ``she", ``mother", ``daughter", ``gal", ``female", ``her", ``herself", ``mary"{]}                                                                                                                                                                                                            \\
	                                                      &
	definitional\_male-Bolukbasi\_et\_al\_2016            &
	{[}``man", ``boy", ``he", ``father", ``son", ``guy", ``male", ``his", ``himself", ``john"{]}                                                                                                                                                                                                                       \\
	                                                      &
	definitional\_female-Bolukbasi\_et\_al\_2016 shuffled &
	{[}"herself", "woman", "daughter", "mary", "her", "girl", "mother", "she", "female", "gal"{]}                                                                                                                                                                                                                      \\
	                                                      &
	definitional\_male-Bolukbasi\_et\_al\_2016 shuffled   &
	{[} "man", "his", "he", "son", "guy", "himself", "father", "boy", "male", "john"{]}                                                                                                                                                                                                                                \\
	                                                      &
	upperclass-Kozlowski\_et\_al\_2019                    &
	{[}``rich", ``richer", ``richest", ``affluence", ``affluent", ``expensive", ``luxury", ``opulent"{]}                                                                                                                                                                                                               \\
	                                                      &
	lowerclass-Kozlowski\_et\_al\_2019                    &
	{[}``poor", ``poorer", ``poorest", ``poverty", ``impoverished", ``inexpensive", ``cheap", ``needy"{]}                                                                                                                                                                                                              \\
	                                                      &
	upperclass-Kozlowski\_et\_al\_2019 shuffled           &
	{[}"richer", "opulent", "luxury", "affluent", "rich", "affluence", "richest", "expensive" {]}                                                                                                                                                                                                                      \\
	                                                      &
	lowerclass-Kozlowski\_et\_al\_2019 shuffled           &
	{[} "poorer", "impoverished", "poorest", "cheap", "needy", "poverty", "inexpensive", "poor"{]}                                                                                                                                                                                                                     \\
	                                                      &
	names\_chinese-Garg\_et\_al\_2018                     &
	{[}``chung", ``liu", ``wong", ``huang", ``ng", ``hu", ``chu", ``chen", ``lin", ``liang", ``wang", ``wu", ``yang", ``tang", ``chang", ``hong", ``li"{]}                                                                                                                                                             \\
	                                                      &
	names\_hispanic-Garg\_et\_al\_2018                    &
	{[}``ruiz", ``alvarez", ``vargas", ``castillo", ``gomez", ``soto", ``gonzalez", ``sanchez", ``rivera", ``mendoza", ``martinez", ``torres", ``rodriguez", ``perez", ``lopez", ``medina", ``diaz", ``garcia", ``castro", ``cruz"{]}                                                                                  \\
	                                                      &
	names\_chinese-Garg\_et\_al\_2018 shuffled            &
	{[}"tang", "chang", "chu", "yang", "wu","hong", "huang", "wong", "hu", "liu", "lin", "chen", "liang", "chung", "li", "ng", "wang"{]}                                                                                                                                                                               \\
	                                                      &
	names\_hispanic-Garg\_et\_al\_2018 shuffled           &
	{[}"ruiz", "rodriguez", "diaz", "perez", "lopez", "vargas", "alvarez", "garcia","cruz", "torres", "gonzalez", "soto", "martinez", "medina", "rivera", "castillo", "castro", "mendoza", "sanchez", "gomez"{]}                                                                                                       \\ \midrule
	\multirow{6}{*}{Fig. \ref{fig:repro_4}}               &
	definitional\_female-Bolukbasi\_et\_al\_2016          &
	{[}``woman", ``girl", ``she", ``mother", ``daughter", ``gal", ``female", ``her", ``herself", ``mary"{]}                                                                                                                                                                                                            \\
	                                                      &
	definitional\_male-Bolukbasi\_et\_al\_2016            &
	{[}``man", ``boy", ``he", ``father", ``son", ``guy", ``male", ``his", ``himself", ``john"{]}                                                                                                                                                                                                                       \\
	{Fig. \ref{fig:repro_4_3}}                            &
	definitional\_female-Bolukbasi\_et\_al\_2016 shuffled &
	{[}"female", "she", "woman", "gal", "her", "daughter", "girl", "herself", "mother", "mary"{]}                                                                                                                                                                                                                      \\
	                                                      &
	definitional\_male-Bolukbasi\_et\_al\_2016 shuffled   &
	{[}"john","man", "son","father", "male","himself", "guy","he", "his","boy"{]}                                                                                                                                                                                                                                      \\
	                                                      &
	random seeds 1                                        &
	{[}``essential", ``want", ``suspension", ``talked", ``competitive", ``information", ``hero", ``bat", ``seconds", ``black"{]}                                                                                                                                                                                       \\
	                                                      &
	random seeds 2                                        &
	{[}``derby", ``passed", ``achieve",, ``discussed", ``providing", ``resulted", ``inmates", ``wearing", ``bid", ``rose"{]}                                                                                                                                                                                           \\ \bottomrule
\end{longtable}
\end{document}